\documentclass[10pt,twocolumn]{article}
\usepackage[utf8]{inputenc}
\usepackage[T1]{fontenc}
\usepackage[margin=0.75in]{geometry}
\usepackage{times}
\usepackage{microtype}
\usepackage{graphicx}
\usepackage{subcaption}
\usepackage{float}
\usepackage{booktabs}
\usepackage{natbib}
\usepackage{hyperref}
\usepackage{amsmath,amssymb}
\usepackage{algorithm}
\usepackage{algorithmic}
\usepackage{titlesec}
\titleformat{\section}{\normalfont\large\bfseries\raggedright}{\thesection}{1em}{}
\titleformat{\subsection}{\normalfont\normalsize\bfseries\raggedright}{\thesubsection}{1em}{}
\titlespacing*{\section}{0pt}{3.5ex plus 1ex minus .2ex}{2.3ex plus .2ex}
\titlespacing*{\subsection}{0pt}{3.25ex plus 1ex minus .2ex}{1.5ex plus .2ex}
\makeatletter
\renewcommand{\maketitle}{%
  \twocolumn[%
    \begin{center}
    {\LARGE\bfseries \@title \par}
    \vspace{0.5em}
    {\large \@author \par}
    \vspace{1em}
    \end{center}
  ]
}
\makeatother

\newcommand{\Exp}{\mathbb{E}}

\title{Speculative Acceptance Does Not Decay with Position:\\
       A Full-Length Diagnostic on Long Chain-of-Thought with Independent Drafts}
\author{Zhiming Meng\\[0.3em]
        \texttt{mengzmemail@163.com}}
\date{}

\begin{document}

\maketitle

\begin{abstract}
The claim that speculative-decoding acceptance rate decays as a chain-of-thought (CoT)
grows longer has become a de~facto industry narrative, motivating draft-model
co-design and position-aware training schemes. Nearly all direct evidence
comes from \emph{trained} drafters whose offline training length is far shorter than
inference, and the length-extrapolation account is the explanation that this
literature itself gives; a purely positional law has therefore never been
isolated.
We present the first systematic, full-length measurement of acceptance rate as a
function of generation position on MATH-500 (all 500 problems), under two decoding
modes (greedy and Leviathan sampling), with two \emph{independent} drafts---neither
trained on target-model outputs---a same-family draft
(DeepSeek-R1-Distill-Qwen-1.5B) and a cross-family draft (Qwen2.5-1.5B-Instruct)
against DeepSeek-R1-Distill-Qwen-7B, yielding $\approx$1.78M verification rounds.
Using within-problem paired tests to remove compositional confounds, we find that
after isolating a greedy-decoding repetition pathology, acceptance rate does
\emph{not} decay with position in any of the four configurations within the
supported ranges (greedy 0--8K, sampling 0--12K); on the contrary,
the second half of each trajectory is slightly but significantly \emph{higher}
than the first half, and the cross-family configuration shows a larger effect.
A controlled temperature sweep further shows that the apparent sampling-mode
degradation of high-entropy reflection tokens is a visibility artifact of
argmax agreement, not a true drop of the expected acceptance probability.
Rejections concentrate with high precision on reflection trigger tokens
(e.g.\ \emph{wait, however, check}), an entropy-selective failure that a
lightweight three-arm LoRA trajectory-alignment intervention (uniform,
reflection-marked, entropy-weighted) fails to improve in mean accepted
length (MAL).
Taken together, we supply the positional baseline that independent
draft--target pairs were missing---not a new mechanism, since the
length-extrapolation account is already the stated explanation in the
trained-drafter literature---and we localize the residual failure to
decision-point entropy rather than to position, redirecting draft alignment
toward objectives that directly optimize the acceptance criterion.
\end{abstract}

\section{Introduction}\label{sec:intro}
Speculative decoding (SD) accelerates large-model inference by having a small
draft model propose multiple tokens that the target model verifies in parallel
\citep{leviathan2023fast,chen2023accelerating}, without changing the output
distribution. Its speedup is governed by the mean accepted length (MAL).
With reasoning models such as DeepSeek-R1 \citep{deepseekai2025r1} routinely
producing chains-of-thought of thousands to tens of thousands of tokens, an
emerging deployment narrative holds that \emph{acceptance rate decays as the
chain grows longer}. The most direct empirical support is Test-Time
Speculation, which reports that trained drafters (EAGLE-3, DFlash, PARD)
degrade monotonically with output length, approaching AL~$\approx$~1 (no
speedup) after $\sim$20K tokens \citep{kumar2026tts}. Recent draft designs
likewise presuppose the decay: DSpark addresses intra-block suffix decay with
semi-autoregressive drafting \citep{cheng2026dspark}, and SharpDraft attributes
long-context decay to attention-mass dilution \citep{park2026sharpdraft}.

Crucially, \emph{all of the above evidence comes from trained drafters} whose
offline training sequences are far shorter than inference. Test-Time
Speculation states this account explicitly: acceptance declines because
speculators are trained offline on short sequences but are forced to match the
target on much longer outputs \citep{kumar2026tts}. We therefore do \emph{not}
present that explanation as our contribution. What is missing is the
controlled baseline that makes it testable: the trajectory of an
\emph{independent} draft--target pair, which carries no training-length
mismatch, over a long CoT. No study has measured that object. The only related
short-range evidence comes from tree-based SD on 64-token generations
\citep{abdellatif2026acceptance}, which does not cover long CoT or use
within-problem pairing.
We supply that baseline, and use it to localize where acceptance actually
fails.

\paragraph{Contributions.}
\begin{itemize}
\item \textbf{A positional baseline for independent drafts}---the primary
contribution (Secs.~\ref{sec:setup}--\ref{sec:result1}). The decay reported for
trained drafters admits a length-extrapolation account that its own paper
states \citep{kumar2026tts}; testing that account requires a draft--target pair
with no training-length mismatch, which has not been measured. We provide it:
MATH-500 in full, two decoding modes, two independent draft families,
$\approx$1.78M verification rounds with per-round position, draft tokens,
target predictions, and (in sampling mode) the full draft/target probability
at each position.
\item \textbf{The baseline does not decay} (Sec.~\ref{sec:result1}). After isolating a
greedy-decoding repetition pathology that inflates long-position acceptance
(Sec.~\ref{sec:artifact}, quantified under both problem-weighted and
round-weighted estimators), within-problem paired tests show that
acceptance rate does not decay with position within the well-supported range;
the second half is slightly but significantly higher in all four
configurations, with the cross-family configuration showing a larger effect.
Combined with contribution~1, this is the controlled comparison that the
existing evidence lacks---not a competing explanation of the trained-drafter
result.
\item \textbf{Entropy, not position, is the failure axis}
(Sec.~\ref{sec:localize})---the novel diagnostic content of this paper. We
establish a dissociation between two acceptance statistics: the \emph{visible}
one (argmax agreement, def2) degrades under sampling, while the \emph{true}
expected acceptance probability $\alpha$ does not. Rejections concentrate with
high precision on reflection trigger tokens rather than on late positions. A
controlled temperature sweep separates the two statistics directly.
\item \textbf{Intervention falsification} (Sec.~\ref{sec:intervention}). A
three-arm LoRA trajectory-alignment intervention (uniform / reflection-marked /
entropy-weighted) yields no MAL gain over the untrained control at 500 problems,
pointing the effective objective away from
trajectory fitting and toward directly optimizing the acceptance criterion
\citep{aa2026acceptance,draftopd2026}.
\end{itemize}

\section{Related Work}\label{sec:related}
\paragraph{Speculative decoding.}
SD was introduced by \citet{leviathan2023fast} and \citet{chen2023accelerating};
the acceptance rule is $\min(1, p(x)/q(x))$ with a residual resampling correction,
and output is distribution-exact. For a survey see \citet{xia2024survey}. Draft
sources include independent small models, Medusa-style heads
\citep{cai2024medusa}, feature-level drafters (EAGLE-3) \citep{li2025eagle3},
and parallel drafters (PARD) \citep{an2025pard}.

\paragraph{Acceptance over long sequences.}
\citet{abdellatif2026acceptance} characterize how acceptance varies with task
domain and tree depth for short tree-based generations, and their reported
position ordering (tokens 33--64 above 1--32) agrees in sign with our rise; we
measure position dependence over 16K-token CoT with within-problem pairing and
artifact removal.
\citet{kumar2026tts} report decay
for trained drafters and attribute it to a training--inference length
mismatch; \citet{cheng2026dspark} and \citet{park2026sharpdraft} address the
decay with semi-autoregressive drafting and query scaling, respectively. Our
contribution relative to this line is a measurement, not a mechanism: we
supply the independent-draft baseline that isolates the mismatch account, and
we show where acceptance does fail once position is ruled out.

\paragraph{Draft alignment and entropy.}
SVIP relates draft entropy to draft--target divergence \citep{zhang2025svip};
EASD proposes entropy-aware rejection resampling \citep{easd2025}. Training
side, SCoT performs thought-level drafting with thinking-behavior alignment
\citep{scot2025alignment}, D-PACE reweights the per-position training loss
toward acceptance-limiting positions \citep{dpace2026}, and acceptance-aware
training directly optimizes expected acceptance length \citep{aa2026acceptance};
Draft-OPD replays verification-exposed errors \citep{draftopd2026}. Our
intervention arm (entropy-weighted) is orthogonal, and its negative result
supports the acceptance-criterion-aware direction.

\section{Measurement Setup}\label{sec:setup}
\paragraph{Models.}
Target: DeepSeek-R1-Distill-Qwen-7B. Drafts: DeepSeek-R1-Distill-Qwen-1.5B
(same family) and Qwen2.5-1.5B-Instruct (cross family)
\citep{deepseekai2025r1,yang2024qwen25}. The prompt is built with the target's
tokenizer/chat template; the cross-family draft runs only on the shared token
sequence, mirroring the real deployment shape of cross-family SD.

\paragraph{Speculative decoding.}
$\gamma=5$. \textbf{Greedy}: draft proposes argmax; target verifies by exact
prefix match, emitting its own argmax at the first mismatch. \textbf{Sampling}
(R1 official serving parameters, $T{=}0.6$, top-$p{=}0.95$): draft samples with
temperature+nucleus; target applies Leviathan sampling verification with
acceptance probability $\min(1, p(x)/q(x))$ and residual resampling
$\mathrm{norm}(\max(0, p{-}q))$ at the first rejection. $p$ and $q$ use the
same temperature/nucleus truncation. When the draft vocabulary (151,936) is
smaller than the target's (152,064), $q$ is zero-padded, which is
mathematically equivalent on the residual. MAL $=(\sum_i \mathrm{accepted}_i
+\#\mathrm{rounds})/\#\mathrm{rounds}$, i.e., including the bonus token,
consistent with vLLM/EAGLE-3.

\paragraph{Two acceptance statistics.}
We distinguish two quantities throughout. \emph{def2} is the visible argmax
agreement: the rate at which the draft's proposed token equals the target's
argmax at that position. It is computable in both modes but is \emph{not} an
expectation under the sampling law. \emph{$\alpha$} is the true expected
acceptance probability, $\alpha=\Exp[\min(1,p/q)]$, evaluated per position over
the draft and target distributions $p,q$. Because this requires the full
distributions, we record $\alpha$ only in sampling mode, where every round
stores the draft and target probability at each draft position. Under greedy
decoding verification is a deterministic argmax prefix match, no position-level
$p,q$ is recorded, and $\alpha$ is therefore not defined there; greedy results
are reported in terms of def2 only.

\paragraph{Why the two statistics dissociate.}
They are not two estimators of one quantity. Since
$\alpha=\sum_x\min(p_x,q_x)=1-\mathrm{TV}(p,q)$, $\alpha$ is a \emph{mass}
statistic; def2 is a \emph{ranking} statistic. Let $a=\arg\max_x p_x$. With
unique argmaxes, $\mathrm{def2}=q(a)\mathbf{1}\{\arg\max_x q_x=a\}$, so
whenever the two argmaxes agree,
\[
\max\bigl\{0,\;p(a)-(1-\alpha)\bigr\}\;\le\;\mathrm{def2}\;\le\;p(a)+(1-\alpha),
\]
and def2 is zero otherwise. The bound fixes def2 only up to an interval of
width $2(1-\alpha)$ around the target's top-1 mass $p(a)$: \emph{$\alpha$
constrains def2 but does not determine it}. The dissociation is largest near
ties. For $p=(\tfrac12+\delta,\tfrac12-\delta)$ and $q$ with the two entries
swapped, $\alpha=1-2\delta$ while def2 $=0$ for every $\delta>0$---an
arbitrarily small ranking perturbation destroys def2 at essentially no cost in
expected acceptance.

This reading is what our measurements show. def2 lies below $\alpha$ in all
eight configuration$\times$token-type cells (Figs.~\ref{fig:toktype}
and~\ref{fig:xfamquad}), the direction expected when the draft's top-1 mass
sits below the target's. The temperature sweep moves def2 by $0.081$
($0.917\to0.836$) while $\alpha$ moves by $0.012$ ($0.927\to0.915$):
temperature perturbs rankings, not overlap. And at 500 problems the
$\alpha-\mathrm{def2}$ gap at reflection tokens is $0.163$ (same-family) and
$0.074$ (cross-family), against $0.006$--$0.057$ at number, operator and text
tokens---largest where near-ties are commonest. A diagnostic built on def2 can
therefore report a large ``degradation'' with no change in expected accepted
length, which is exactly the pattern we observe.

\paragraph{Data and instrumentation.}
MATH-500 \citep{hendrycks2021math,lightman2023verify}, all 500 problems,
seed 42, and a budget of 16,384 new tokens per problem. Per round we record
position, the 5 draft tokens, the 6 target predictions, and accepted count;
in sampling mode we additionally record the draft/target probability at each
draft position. We report four configurations, abbreviated \emph{same}
(same-family) and \emph{xfam} (cross-family): same/greedy, same/sample,
xfam/greedy, xfam/sample (394,572 / 382,556 / 499,428 / 502,976 rounds).

\paragraph{Statistics.}
All before--after comparisons are \emph{within-problem paired}: for each
problem, sort rounds by position, split at the median, compute the
second-half minus first-half acceptance difference. Group-level significance
uses an exact two-sided binomial sign test (problems with zero difference
excluded) and problem-level bootstrap (10,000 resamples) 95\% CIs.
We exclude ``loop'' problems---generations that exhaust the 16,384-token
budget---before the paired tests. Operationally the flag is a length-only rule
(total generated length $\ge$16,000 tokens); the exclusion rule uses length
only, never correctness (all such generations happen to have wrong final
answers; Sec.~\ref{sec:artifact}), and the conclusions are insensitive to
whether the flag is set at 16,000 or at the full cap.

\paragraph{Correctness checks.}
We verified the implementation against the original algorithm
(App.~\ref{app:correctness}): per-round $\Exp[\min(1,p/q)]$ matches observed
acceptance to $4\times10^{-4}$; a synthetic-distribution unit test shows the
output distribution is distribution-exact; the tokenizer alignment across
families is verified (0 id conflicts on 151,658 shared tokens); and the four
core data files are bit-identical between local and remote.

\begin{figure*}[!t]
\centering
\includegraphics[width=\textwidth]{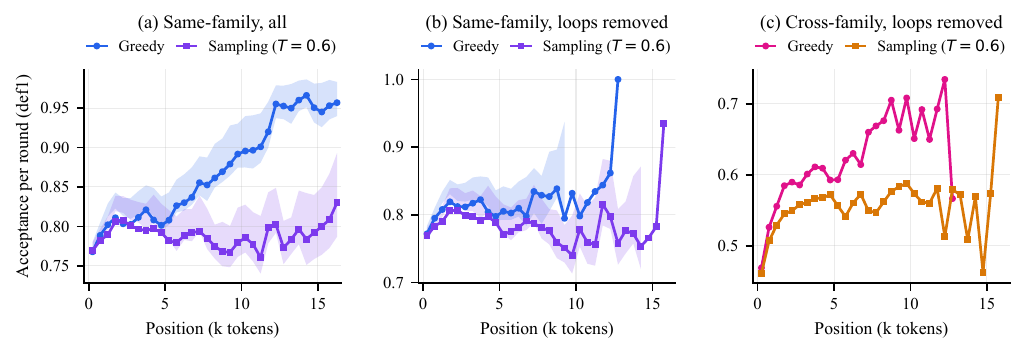}
\caption{Acceptance rate vs.\ generation position. (a) Same-family, all problems:
the greedy curve rises and saturates after 12K---almost entirely driven by loop
problems. (b) Same-family after removing them: no downward trend, and if anything
a slight rise (greedy $0.91$--$0.94$, sampling $0.856$--$0.882$). (c) Cross-family
(loops removed) shows the same shape at a lower level. Effective support: greedy
0--8K, sampling 0--12K; beyond these ranges few problems remain and the curves
become noisy (Sec.~\ref{sec:limitations}), so we do not draw conclusions there.}
\label{fig:pos}
\end{figure*}

\section{Acceptance Does Not Decay (and Rises Slightly)}\label{sec:result1}
Fig.~\ref{fig:pos} shows the main result. Two observations. First, \emph{neither
mode shows decay} within the well-supported range (greedy 0--8K, sampling
0--12K): after loop removal, greedy def2 fluctuates in $0.91$--$0.94$ and
sampling def2 in $0.856$--$0.882$, with no downward trend. Second, the greedy
full curve's ``rise + right-tail saturation'' ($0.78 \to 0.95+$) is largely an
artifact of loop problems; after removal it reduces to a mild but significant
within-problem rise.

\begin{table}[t]
\centering\small
\caption{Within-problem paired half-difference (second $-$ first) after loop
removal. def1 $=\mathbb{E}[\mathrm{accepted}]/\gamma$;
$\alpha=\Exp[\min(1,p/q)]$ is the true expected acceptance probability; it is
reported only for sampling configurations because greedy decoding does not
record position-level $p,q$ (Sec.~\ref{sec:setup}). 95\% CIs are problem-level
bootstrap (10,000 resamples);
``Up/Down'' counts problems whose second-half acceptance exceeds / falls below
the first half (zero differences excluded); $p$ is the exact two-sided sign
test. def2 (argmax-match rate) curves appear in Fig.~\ref{fig:pos} and are
excluded from paired tests because def2 is not an expectation over the
sampling law (Sec.~\ref{sec:localize}).}
\label{tab:halfdiff}
\resizebox{\columnwidth}{!}{%
\begin{tabular}{llcccc}
\toprule
Config & Metric & Diff & 95\% CI & Up/Down & $p$ \\
\midrule
same/greedy  & def1 & $+0.031$ & $[+0.026,+0.036]$ & 327/126 & $1.2{\times}10^{-21}$ \\
same/sample  & def1 & $+0.028$ & $[+0.023,+0.033]$ & 354/130 & $5.2{\times}10^{-25}$ \\
xfam/greedy  & def1 & $+0.069$ & $[+0.063,+0.076]$ & 393/61  & $1.8{\times}10^{-60}$ \\
xfam/sample  & def1 & $+0.059$ & $[+0.053,+0.065]$ & 407/78  & $9.6{\times}10^{-55}$ \\
\midrule
same/sample  & $\alpha$ & $+0.010$ & $[+0.008,+0.013]$ & 333/151 & $8.3{\times}10^{-17}$ \\
xfam/sample  & $\alpha$ & $+0.030$ & $[+0.027,+0.034]$ & 365/120 & $8.7{\times}10^{-30}$ \\
\bottomrule
\end{tabular}}
\end{table}

Table~\ref{tab:halfdiff} gives the paired tests. \emph{All four configurations
show a small but significant rise}, with the cross-family configuration roughly
doubling the effect ($+0.069/+0.059$). Crucially, the sampling-mode $\alpha$
(a distribution-level quantity) also rises, ruling out the confound that def2
rises merely because the target distribution becomes sharper with position.
The same pattern holds when each trajectory is split into within-problem
position quintiles instead of halves (Fig.~\ref{fig:quintiles}): acceptance is
flat to slightly rising from Q1 to Q5 in both modes. The sign of the effect is
independently corroborated at short range: \citet{abdellatif2026acceptance}
report higher acceptance at tokens 33--64 than at 1--32 in a tree-based
64-token setting, and our curves extend that ordering from 64 tokens to the
8--12K range.

\paragraph{Scope.}
We do not claim decay is impossible. We claim that \emph{unconditional,
positional} decay is absent on MATH-500 CoT for these model pairs, within
greedy 0--8K and sampling 0--12K effective support. This does not contradict
trained-drafter decay \citep{kumar2026tts}: that literature attributes it to a
training--inference length mismatch, and a non-decaying independent-draft curve is
exactly what the mismatch account predicts once the mismatch is removed. Our
contribution on this axis is the baseline that makes the prediction testable,
not the explanation itself; Sec.~\ref{sec:localize} supplies the part of the
diagnosis that is not already implied by that account.

\section{A Repetition Artifact Inflates Greedy Tail Acceptance}\label{sec:artifact}
Greedy R1-Distill has a known pathology: a minority of problems fall into an
``infinite self-reverification spiral'' that exhausts the 16,384-token budget.
We flag \emph{loop problems} by a length-only rule---total generated length
$\ge$16,000 tokens. Greedy has 46 such problems (9.2\%), sampling 16 (3.2\%),
with 8 overlapping; flagging instead at the full 16,384-token cap leaves the
greedy count unchanged at 46 and reduces the sampling count to 15. Every
flagged generation happens to have a wrong final answer,
but correctness is never used as a criterion, so the exclusion cannot
condition on the outcome. The rule is also not sensitive to the threshold:
varying it from 8K to 16K tokens changes the four paired differences of
Table~\ref{tab:halfdiff} by less than $0.001$, and moving from 16,000 to the
16,384-token cap changes them by less than $10^{-4}$.

\begin{figure}[t]
\centering
\includegraphics[width=\columnwidth]{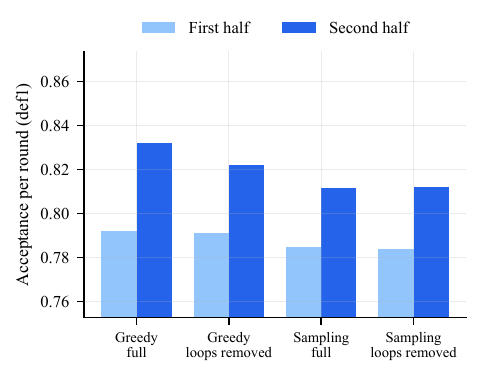}
\caption{First-half vs.\ second-half acceptance (def1), before and after loop
removal. Bars use the \emph{problem-weighted} paired estimator of
Table~\ref{tab:halfdiff}: each problem contributes one half-pair, so the 46
loop problems have bounded influence and the greedy excess shrinks only
modestly after removal ($+0.040 \to +0.031$; sampling $+0.027 \to +0.028$).
The dominant loop effect is a \emph{round-weighting} phenomenon: in the
pooled statistics (Table~\ref{tab:artifact}: $+0.064 \to +0.025$) loop rounds
dominate long positions, and the position curves flatten accordingly
(Fig.~\ref{fig:pos}(a) vs.\ (b)).}
\label{fig:artifact}
\end{figure}

These problems contribute most of the samples above 12K and behave
programmatically---40-char-overlap shingle checks show sentence-level
self-reverification rather than verbatim loops (verbatim repeat rate $\approx0$;
5/46 greedy problems have tail verbatim repeat $0.43$--$0.50$). Their anomalous
near-$0.99$ acceptance pollutes long-position statistics. The two estimators
bracket the artifact (Fig.~\ref{fig:artifact}, Table~\ref{tab:artifact}): the
\emph{round-weighted} pooled half-difference drops by $\sim$60\% after loop
removal in the same-family greedy configuration ($\sim$50\% cross-family),
whereas the \emph{problem-weighted} paired difference drops only from $+0.040$
to $+0.031$---loop problems are 46 of 500, so once every problem counts equally
their influence is bounded, and the artifact is primarily a round-weighting
effect on long-position statistics. A mild significant rise remains in the
paired estimator (Table~\ref{tab:halfdiff}), which App.~\ref{app:oaxaca}
decomposes.

\begin{table}[t]
\centering\small
\caption{Aggregate half-difference before/after loop removal. Unlike
Table~\ref{tab:halfdiff}, rounds are \emph{pooled across problems} and split
at the global median position, so loop problems (which contribute most
long-position rounds) dominate the pre-removal column; the paired column of
Table~\ref{tab:halfdiff} weights every problem equally and is the primary
statistic.}
\label{tab:artifact}
\begin{tabular}{lcc}
\toprule
Config & Full & Loop removed \\
\midrule
same/greedy  & $+0.064$ & $+0.025$ \\
same/sample  & $+0.006$ & $+0.009$ \\
xfam/greedy  & $+0.172$ & $+0.090$ \\
xfam/sample  & $+0.060$ & $+0.056$ \\
\bottomrule
\end{tabular}
\end{table}

\section{Rejections Concentrate on Reflection Tokens}\label{sec:localize}
\begin{figure*}[!t]
\centering
\includegraphics[width=\textwidth]{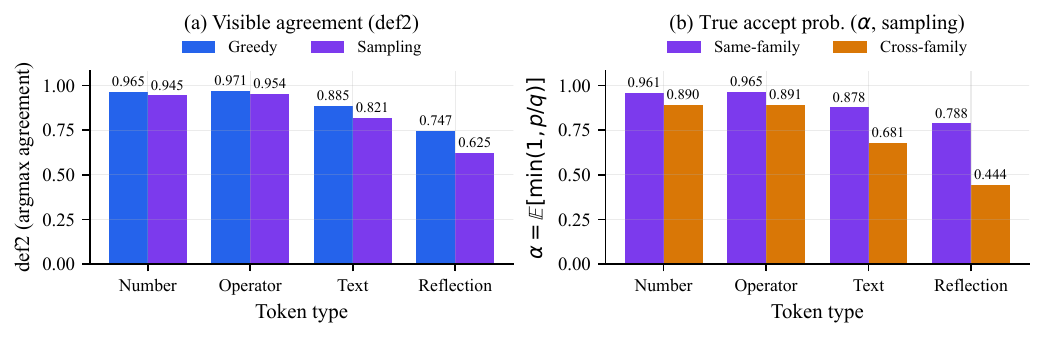}
\caption{def2 (visible argmax agreement, both modes) and $\alpha$ (true expected
acceptance probability, sampling only---greedy records no position-level $p,q$)
by token type, non-loop. Failure is entropy-selective: number/operator
are near-perfect, reflection tokens are the clear failure point, and cross-family
misalignment amplifies it.}\label{fig:toktype}
\end{figure*}

\begin{figure}[t]
\centering
\includegraphics[width=\columnwidth]{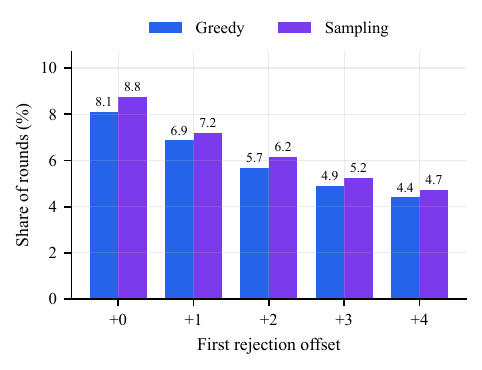}
\caption{Mismatch structure at reflection triggers.}\label{fig:mismatch}
\end{figure}

Classifying draft tokens into number / operator / text / reflection trigger
(\emph{wait, however, check, verify, \dots}) shows that failure is
entropy-selective (Fig.~\ref{fig:toktype}): number/operator tokens are accepted
at $\ge 0.93$ in all configurations, while reflection tokens are the clear
failure point. Under greedy, non-loop reflection def2 is $0.747$, and
cross-family misalignment compresses it to $0.367$. In sampling mode the same
bucket has $\alpha=0.788$, which falls to $0.444$ under cross-family
misalignment (def2 $0.625\to0.369$). Fig.~\ref{fig:mismatch} locates these
rejections inside the draft block, showing the offset at which the first
rejection occurs.

\paragraph{Sampling changes visibility, not acceptance.}
Sampling makes the high-entropy failure \emph{visible}: reflection def2 drops
$0.747 \to 0.625$ on the full 500-problem non-loop sample (the 100-problem
temperature-sweep subset of Table~\ref{tab:temp} gives $0.636$; the two numbers
differ only in sample size), but the true expected acceptance probability
$\alpha$ does \emph{not} drop ($0.788 \ge 0.747$), consistent with the overall
MAL cost of sampling being only $3.4\%$. The intuition is that argmax agreement is a poor
proxy at high entropy: even a perfectly aligned draft would rarely hit the
target's argmax once the distribution flattens, while the $\min(1,p/q)$ rule
rejects at the true distribution level. The bound of Sec.~\ref{sec:setup}
makes this quantitative: def2 is pinned to the target's top-1 mass only within
$1-\alpha$ of mass overlap, so a spread-out draft loses ranking agreement
without losing acceptance mass.

\paragraph{Composition shift.}
Long CoTs drift toward more procedural content (numbers, operators) in the
second half, which mechanically raises acceptance. An Oaxaca--Blinder
decomposition \citep{oaxaca1973,blinder1973} attributes $30$--$45\%$ of the
residual rise to this composition shift (App.~\ref{app:oaxaca}); the rest is a
genuine within-problem effect.

\begin{figure}[t]
\centering
\includegraphics[width=\columnwidth]{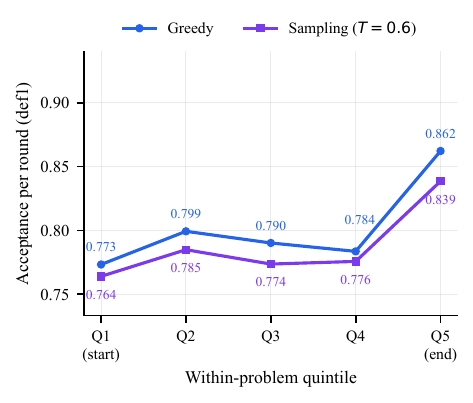}
\caption{Within-problem position quintiles: acceptance is flat-to-rising after
loop removal.}\label{fig:quintiles}
\end{figure}

\begin{table}[t]
\centering\small
\caption{Temperature sweep (same-family sampling, first 100 problems, loop
removed). def2 falls sharply with $T$, but the true expected acceptance
probability $\alpha$ barely moves---the apparent degradation of reflection
tokens is a visibility artifact.}
\label{tab:temp}
\resizebox{\columnwidth}{!}{%
\begin{tabular}{lccccc}
\toprule
$T$ & MAL & def2 & $\alpha$ & reflection def2 & reflection $\alpha$ \\
\midrule
0.3 & 4.986 & 0.917 & 0.927 & 0.707 & 0.785 \\
0.6 & 4.967 & 0.886 & 0.922 & 0.636 & 0.802 \\
0.9 & 4.884 & 0.836 & 0.915 & 0.520 & 0.809 \\
\bottomrule
\end{tabular}}
\end{table}

\paragraph{Temperature sweep.}
We sweep $T\in\{0.3,0.6,0.9\}$ (Table~\ref{tab:temp}). def2 falls sharply with
$T$ ($0.917\to0.836$), and reflection def2 collapses ($0.707\to0.520$), yet
$\alpha$ barely moves ($0.927\to0.915$) and reflection $\alpha$ slightly
\emph{rises} ($0.785\to0.809$). This confirms that the sampling-mode
``degradation'' of reflection tokens is a visibility artifact of argmax
agreement, not a true drop of the acceptance probability.

\section[Entropy-Targeted Draft Training Does Not Help]{Entropy-Targeted Draft Training\\Does Not Help}\label{sec:intervention}
\begin{figure}[t]
\centering
\includegraphics[width=\columnwidth]{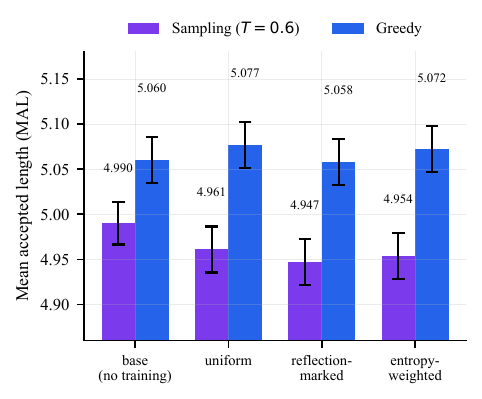}
\caption{MAL at 500 problems under the three intervention arms and the
untrained base draft (same-family; error bars: standard error). No arm
improves over the uniform training control in either decoding mode.}\label{fig:mal}
\end{figure}

The diagnosis---failure concentrated at high-entropy reflection decision
points, entropy-selective, amplified by family misalignment---naturally
suggests that trajectory-alignment training targeted at high-entropy positions
should repair it. We test the most direct form with a three-arm LoRA ($r{=}32$)
intervention: \textbf{uniform} (all positions), \textbf{reflection-marked}
(weighted toward reflection triggers), \textbf{entropy-weighted} (weighted by
position entropy). Training data: 15,000 MATH training-set trajectories
(disjoint from MATH-500), correctness-filtered to 10,279, with per-position
entropy annotation from both roles (1.82M tokens).

\begin{table}[t]
\centering\small
\caption{Intervention results at 500 problems (same-family, both decoding
modes). MAL is reported with standard errors. $\Delta$ columns are
problem-paired differences against the \emph{uniform} arm---the proper
control, since it undergoes identical training on the same trajectories with
flat weighting---computed from unrounded per-problem MAL, so a $\Delta$ need
not equal the difference of the two rounded column entries; 95\% CIs are from
paired $t$-tests. No arm improves over the uniform control; answer accuracy is
identical across arms (McNemar $\chi^2{\approx}0$).}
\label{tab:intervention}
\resizebox{\columnwidth}{!}{%
\begin{tabular}{lccccc}
\toprule
Arm & MAL$_{\mathrm{samp}}$ & MAL$_{\mathrm{greedy}}$ & $\Delta_{\mathrm{samp}}$ [95\% CI] & $\Delta_{\mathrm{greedy}}$ [95\% CI] \\
\midrule
base (no training)  & $4.990{\pm}.012$ & $5.060{\pm}.013$ & -- & -- \\
uniform             & $4.961{\pm}.013$ & $5.077{\pm}.013$ & -- & -- \\
reflection-marked   & $4.947{\pm}.013$ & $5.058{\pm}.013$ & $-0.0138$ $[-0.032,+0.005]$ & $-0.0192$ $[-0.042,+0.003]$ \\
entropy-weighted    & $4.954{\pm}.013$ & $5.072{\pm}.013$ & $-0.0072$ $[-0.026,+0.012]$ & $-0.0055$ $[-0.008,-0.003]$ \\
\bottomrule
\end{tabular}}
\end{table}

\emph{No arm improves MAL over the uniform training control}
(Table~\ref{tab:intervention}). Fig.~\ref{fig:mal} shows the four arms in both
modes; every interval overlaps the uniform control, and the sampling and greedy
orderings are not even consistent with each other. All paired differences are
at most $0.02$
tokens per round ($<0.4\%$), and every confidence interval includes or
borderline-includes zero; the single nominally significant cell (greedy,
entropy $-$ uniform $= -0.0055$, $t{=}-4.14$) corresponds to five and a half
thousandths of a token with effect size $dz{=}-0.19$, and its sign is
\emph{negative}---the trained arm is, if anything, marginally worse. Scaling from the 100-problem
pilot to all 500 problems sharpens rather than weakens the null: per-arm
standard errors fall from $\approx$0.032 to $\approx$0.013 (a factor of 2.4,
matching $\sqrt{500/100}$), while the between-arm differences fall in step, so
this is a confirmed null effect, not an underpowered one.\footnote{Pilot
($n{=}100$) MAL mean$\pm$SE: uniform $4.928{\pm}0.030$ (sampling) and
$5.048{\pm}0.032$ (greedy); reflection-marked $4.922{\pm}0.032$ and
$5.042{\pm}0.033$; entropy-weighted $4.920{\pm}0.031$ and $5.044{\pm}0.032$.
All pilot arms lie within one pilot SE of each other, so the pilot already
showed no separation; the 500-problem run narrows the interval that excludes
any effect larger than $\approx$0.03 tokens.} Accuracy is
identical across arms (McNemar $\chi^2{\approx}0$), ruling out a
training-damage account of the null. Meanwhile, on the 100-problem diagnostic
subset, argmax agreement at reflection tokens rises by $6.4$ percentage points
over the untrained draft in the uniform arm ($0.645\to0.709$) and by $6.3$ in
the entropy-weighted arm, while the reflection-marked arm lowers it to
$0.551$ (Fig.~\ref{fig:w3type})---yet even the largest of these visible
alignment gains does not convert into MAL, mirroring the def2/$\alpha$
dissociation of Sec.~\ref{sec:localize}.

Why does the marker-weighted arm move \emph{away} from the target on exactly the
tokens it upweights? The reweighting is narrow: markers occupy $0.85\%$ of
response positions (measured on the same-family sampling traces), and after the
per-trajectory mean-1 normalization they carry $\approx4\times$ the weight of an
ordinary position while all other positions are scaled by $0.97$. The marked set
is a closed list of eight single-token strings, so the extra gradient repeats
over a small, highly context-dependent set of positions---the regime in which a
rank-32 adapter trained for two epochs fits context-specific marker choices
instead of generalizing. Two checks argue against a broken run: agreement on
text and number tokens stays within $0.0002$ of the uniform arm ($0.8436$ vs.\
$0.8438$; $0.9506$ for both), which a corrupted adapter would not leave intact;
and the effective sample size of the reweighted objective,
$(1+3p)^2/(1+15p)\approx0.93$ at $p{=}0.0085$, is only $7\%$ below the uniform
arm's, so a collapse of effective sample size is ruled out. What remains is a
shift in the draft's marker marginal---it proposes markers in contexts where the
target does not. We therefore read this arm as a diagnostic of concentrated
reweighting rather than as evidence about whether trajectory fitting can help:
the causal contrast is uniform vs.\ entropy, which share trajectories, schedule,
and step count and differ only in weight dispersion.

The negative result is informative in a narrower sense than a mechanism claim:
a draft that is explicitly trained to match the target on these very
trajectories still does not improve MAL, so the residual rejection at
reflection points is not a draft \emph{knowledge} deficit that SFT-style
trajectory fitting can repair. A natural reading is that these decision points
are intrinsically high-variance for the target, but what the experiment
establishes is the negative constraint---trajectory fitting does not help---not
the positive mechanism. Effective draft alignment must therefore move from
``fit the target trajectory'' to ``optimize the acceptance criterion''
(e.g.\ $\Exp[\min(1,p/q)]$ or a proxy), as in acceptance-aware training
\citep{aa2026acceptance} and Draft-OPD \citep{draftopd2026}; our diagnostic
protocol and negative evidence provide a quantitative basis for this turn.

\section{Discussion and Limitations}\label{sec:limitations}
\paragraph{Cross-family as a diagnostic amplifier.}
The cross-family draft is a ``misalignment amplifier'': it stretches the
same-family repair headroom from $+0.3\%$ to a theoretical MAL gain of $+25\%$.
However, its $\gamma{=}5$ throughput ($24.3$ tok/s) is already below the
target's pure autoregressive throughput ($46.3$ tok/s), so its value is
diagnostic, not deployable. All cross-arm throughput and wall-clock numbers
reported here were collected on a single physical GPU; App.~\ref{app:hardware}
documents the multi-GPU audit and the paired hardware-control run.

\paragraph{Limitations.}
(i) Support is limited at long positions: after loop removal, greedy has 15
problems at 8K and 4 at 9K; sampling has 33 at 8K and 11 at 12K---we restrict
strong claims to those ranges. (ii) The reflection bucket is small
($1.6$--$2$K tokens) in the temperature sweep, so the $\alpha$ rise there is
suggestive, not definitive. (iii) A single target family, a single seed, and a
fixed $\gamma{=}5$ bound external validity; a Qwen3-8B/1.7B replication with
paired cross-family acceptance logging, multi-seed runs, and a $\gamma$ sweep
are future work. (iv) We do not train a length-extrapolating drafter, so what
we report is the independent-draft baseline rather than a direct test of the
training--inference mismatch account. Running a trained drafter (EAGLE-3,
PARD, or DFlash) under the same full-length protocol is the experiment that
would close that loop, and it is the single most important extension of this
work. The bound of Sec.~\ref{sec:setup} makes that extension a sharp test: a
drafter trained on target outputs should over-concentrate on the target's
top token, which pushes def2 toward $\alpha$ and can push it above, whereas
the flat $\alpha$ profile we measure implies that any decay such a drafter
shows must come from mass (TV) rather than from ranking. Logging the target
top-1 mass $p(a)$ alongside $p,q$ would adjudicate the two readings directly.
(v) MATH-500 is a single domain; extending
to other long-form domains is future work.

\section{Conclusion}\label{sec:conclusion}
We provide the first full-length, controlled measurement of speculative
acceptance over position for \emph{independent} draft--target pairs on long
CoT---the baseline against which trained-drafter decay should be read, not a
competing explanation of it. After isolating a greedy-decoding repetition
artifact and using within-problem pairing, acceptance does not decay with
position in any configuration within the supported ranges (greedy 0--8K,
sampling 0--12K); if anything it rises slightly. With position thus ruled out
as an axis of degradation, we localize the failure on the axis that does
matter: rejections concentrate on high-entropy
reflection decision points, where the visible (argmax) and true ($\alpha$)
acceptance statistics dissociate. Trajectory alignment cannot repair this,
which points the objective of draft alignment at the acceptance criterion
itself rather than at trajectory fitting. We hope the protocol and the
negative result redirect a growing line of draft-alignment work toward
acceptance-aware objectives.

\bibliographystyle{plainnat}
\bibliography{references}

\clearpage
\appendix
\onecolumn
\section{Reproducibility and Correctness of the Measurement}\label{app:correctness}
Position conclusions are meaningful only if the measurement pipeline is
correct. We therefore validate it in two layers.

\paragraph{Algorithmic fidelity.}
We align the implementation step-by-step against the original Algorithm~1 of
\citet{leviathan2023fast}, cross-checked against \citet{chen2023accelerating}
and the vLLM EAGLE-3 implementation; all steps match. Four measurement
conventions differ from a literal reading of the algorithm and are quantified:
(i) temperature+top-$p$ truncation makes exactness hold for the truncated
target distribution $\tilde p$ (the same convention as standard inference
frameworks; our autoregressive baseline uses the same $T$/top-$p$);
(ii) the recorded position is the round-start confirmed length, a constant
$+1$ offset that does not change any binning or half-split conclusion
(recomputing with true positions changes the four paired differences by
$<0.002$); (iii) EOS inside the accepted prefix truncates the output text but
still counts the round, affecting $0.06\%$ of tokens; (iv) a pending bonus
round runs one extra draft forward (throughput only, not distribution).

\paragraph{Data integrity and internal consistency.}
(a) The measurement, configuration, and analysis code, and all eight core
data files, are bit-identical between the local workstation and the remote
server (verified by cryptographic hash). (b) All four datasets parse cleanly
(2,000 problems, 1,779,532 rounds), with zero malformed lines, zero duplicate
keys, and zero violations of the structural invariants (round continuity,
$\gamma{=}5$, accepted-count range, and the MAL identity). (c) The per-round
$\Exp[\min(1,p/q)]$ computed from the recorded $p,q$ matches observed
acceptance to $4\times10^{-4}$ (expected vs.\ observed accepted per round:
same-family $3.9447$ vs.\ $3.9441$; cross-family $2.6694$ vs.\ $2.6703$;
per-position survival max deviation $4\times10^{-4}$). (d) A synthetic
distribution unit test shows the speculative sampler is distribution-exact
(total variation $0.00375$ vs.\ direct sampling $0.00346$; max deviation
$0.00079$, below the sampling noise floor $0.00158$). (e) Tokenizer alignment
across families is verified (0 id conflicts on 151,658 shared tokens; the
cross-family draft adds only 7 tokens). All checks were re-run at a second,
independent time point with identical results.

\section{Hardware and Throughput Measurement Policy}\label{app:hardware}
Every throughput (tok/s) and wall-clock comparison reported in this paper was
collected on a single physical GPU per comparison. This policy matters because
the two devices in our pool---an RTX 4090 and an RTX 3090 Ti---differ by
5--15\% on batch-1 decoding, enough to masquerade as a method effect. Two
properties make the paper's conclusions immune even where a configuration was
split across devices. First, MAL is computed from the recorded acceptance path
and, with a fixed seed and prompts, is a deterministic function of the model
pair; device speed cannot enter it. Second, answer-match is a property of the
generated text alone. Consequently, every number in
Tables~\ref{tab:halfdiff}--\ref{tab:intervention} is either a MAL-type or an
accuracy-type quantity, and no throughput number from a multi-device
configuration is used anywhere in this paper.

An audit of the intervention runs found one early configuration that violated
the single-device policy: the three arms had been scheduled as
reflection-marked on the 4090 and uniform/entropy-weighted on the 3090 Ti. The
resulting apparent throughput advantage of the reflection-marked arm
($+5.0\%$ sampling, $+3.6\%$ greedy) is fully consistent with that split and is
therefore withdrawn; it is not reported as a method effect and does not appear
in any table. The MAL null result is unaffected, both because MAL is
device-invariant by construction and because the sign of the MAL differences is
\emph{opposite} to the throughput ordering, which excludes a hardware
explanation of the null.

As a determinism check, the uniform checkpoint was re-run under the pilot
protocol (100 problems, seed 42, $T{=}0.6$, top-$p{=}0.95$; 81,347 rounds) and
reproduced the pilot's MAL to four decimals ($4.9284\pm0.0302$), confirming
that the acceptance path is reproducible across runs. A paired two-device run
of the same checkpoint is the direct way to quantify the residual hardware
offset; because no cross-device throughput claim is made in this paper, it does
not affect any reported number.

\section{Oaxaca--Blinder Decomposition of the Residual Rise}\label{app:oaxaca}
\begin{figure}[H]
\centering
\begin{minipage}[t]{0.49\textwidth}
\centering
\includegraphics[width=\textwidth]{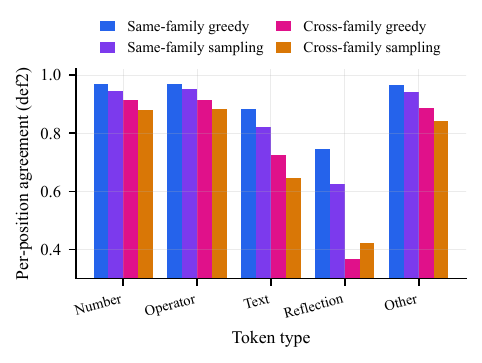}
\end{minipage}\hfill
\begin{minipage}[t]{0.49\textwidth}
\centering
\includegraphics[width=\textwidth]{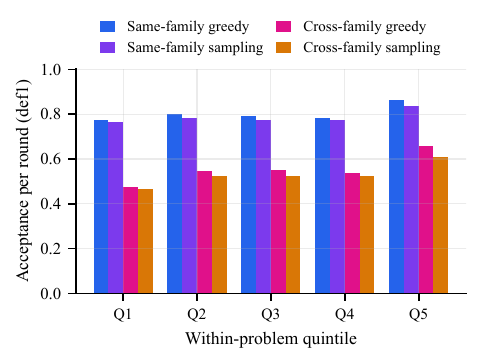}
\end{minipage}
\caption{Cross-family counterparts of Figs.~\ref{fig:toktype}
and~\ref{fig:quintiles}: (left) token-type breakdown and (right)
within-problem position quintiles.}\label{fig:xfamquad}
\end{figure}

We decompose the within-problem second-half $-$ first-half acceptance
difference into a composition component (change in token-type mix) and a
rate component (change in per-type acceptance), using the Oaxaca--Blinder
decomposition \citep{oaxaca1973,blinder1973}. Across the four configurations,
$30$--$45\%$ of the residual rise is attributed to composition shift (long CoTs
become more procedural in the second half), leaving a genuine within-problem
rate effect. Full per-configuration tables are in the code release.
Fig.~\ref{fig:xfamquad} gives the cross-family
counterparts of Figs.~\ref{fig:toktype} and~\ref{fig:quintiles}.

\section{Additional Figures}
\begin{figure}[H]
\centering
\includegraphics[width=0.52\textwidth]{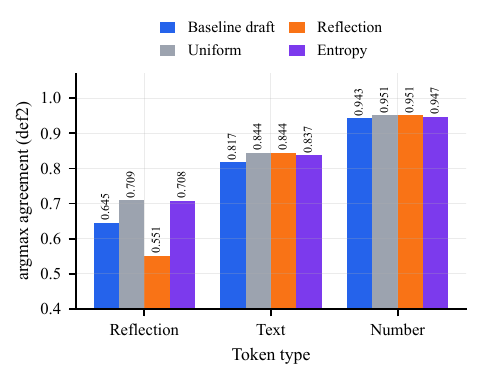}
\caption{Token-type breakdown for the intervention experiment (100-problem
diagnostic subset).}\label{fig:w3type}
\end{figure}

\end{document}